\documentclass[runningheads]{llncs}
\usepackage[T1]{fontenc}
\usepackage{newtxtext}       %
\usepackage[varvw]{newtxmath}       

\usepackage{graphicx}

\usepackage{amsmath}
\usepackage{cite}
\usepackage{orcidlink}

\newcommand{\R}{\ensuremath{\mathbb R}}

\newcommand*{\E}{\ensuremath{\mathbb E}}

\newcommand{\bfx}{\ensuremath{\mathbf{x}}}
\newcommand{\bfy}{\ensuremath{\mathbf{y}}}
\newcommand{\bfw}{\ensuremath{\mathbf{w}}}
\newcommand{\bfm}{\ensuremath{\mathbf{m}}}
\newcommand{\bfdelta}{\ensuremath{\boldsymbol{\delta}}}
\begin{document}
\title{Deep Evidential Regression for Sparse Forest Height Estimation from Multimodal Satellite Imagery}
\titlerunning{DER for Sparse Forest Height Estimation}
%
\author{Laura Bader \inst{1,3}\orcidlink{0009-0008-9756-3462} \and
Muhammad Ammar Ahmed\inst{2}\orcidlink{0009-0009-8246-0599} \and
Xiao Xiang Zhu \inst{2,3}\orcidlink{0000-0001-5530-3613} \and
Göran Kauermann \inst{1,3}\orcidlink{0000-0003-0742-7835}}
\authorrunning{L. Bader et al.}
%
\institute{LMU Munich, 80539 Munich, Germany \\ \email{laura.bader@lmu.de}\\ \and
TU Munich, 80333 Munich \\ \and
Munich Center for Machine Learning (MCML)
}

\maketitle              
\begin{abstract}
Accurate estimation of forest height from satellite imagery is essential for applications such as carbon accounting, biodiversity monitoring, and ecosystem management. While recent deep learning approaches provide accurate predictions, they typically do not quantify predictive uncertainty. This limitation is particularly relevant in geospatial settings characterized by sparse supervision and geographic distribution shift.
In this work, we investigate Deep Evidential Regression (DER) for forest height estimation on the TreeUQ benchmark, a large-scale dataset designed for the joint estimation of tree count and average tree height at 10 m resolution, based on Sentinel-1/-2 data as well as tree inventory data over the federal state of Bavaria. To account for the extreme label sparsity of the tree inventory data, we introduce a masked evidential loss for dense geospatial prediction. Using a U-Net architecture with multimodal Sentinel-1 and Sentinel-2 inputs, the proposed approach jointly predicts tree height and associated uncertainty estimates in a single forward pass.
Experimental results show that DER achieves predictive performance comparable to a deterministic U-Net while additionally providing well-calibrated uncertainty estimates. These findings demonstrate the potential of evidential learning as an efficient framework for uncertainty-aware forest structure estimation from Earth observation data.

\end{abstract}

\section{Introduction}
Exact estimation of forest structure from satellite imagery is essential for applications in climate science, biodiversity monitoring, and ecosystem management. Recent advances in multimodal Earth observation have enabled large-scale prediction of forest attributes such as canopy height using Sentinel-1 SAR and Sentinel-2 optical imagery \cite{lang_high-resolution_2023,potapov_mapping_2021}. However, most existing approaches focus exclusively on deterministic point predictions and do not quantify predictive uncertainty, despite uncertainty being critical for reliable ecological decision making \cite{kujala_role_2023}.
Estimating tree heights from satellite imagery is inherently uncertain due to sensor noise, heterogeneous vegetation structure, and geographic distribution shift. These challenges are amplified in sparse supervision settings, where only a small subset of pixels contains valid annotations. The recently introduced TreeUQ benchmark \cite{ahmed_treeuq_2026} highlights this problem by providing multimodal satellite imagery together with sparse tree inventory labels for forest structure estimation at 10 m resolution for the region of Bavaria in Germany. While TreeUQ explicitly motivates uncertainty-aware modeling, existing baseline methods remain fully deterministic.
In this work, we investigate Deep Evidential Regression (DER) for uncertainty-aware forest height estimation from multimodal satellite imagery. DER estimates predictive uncertainty in a single forward pass by directly parameterizing an evidential distribution over regression outputs, avoiding repeated stochastic inference or model ensembles. To apply DER in the sparse supervision setting of TreeUQ, we introduce a masked evidential training objective that restricts learning to pixels with valid tree inventory annotations. Using a U-Net backbone and multimodal Sentinel-1 and Sentinel-2 inputs, the proposed approach jointly predicts tree height together with aleatoric and epistemic uncertainty estimates.

More precisely, our contributions are threefold: (1) We adapt DER to sparse multimodal Earth observation regression for forest height estimation. (2) We introduce a masked evidential loss formulation that enables DER model training under extreme geospatial label sparsity. (3) We provide a comprehensive empirical evaluation of predictive uncertainty, including calibration analysis, spatial uncertainty characterization, and an investigation of the relationship between uncertainty, forest structural heterogeneity, and prediction error. 

The code is made available at \url{https://github.com/ammarlam10/evidential}.

\section{Related work}
Uncertainty quantification has become an important topic in deep learning, particularly in safety-critical and scientific applications \cite{gawlikowski_survey_2023}. Common approaches include Bayesian neural networks, Monte Carlo dropout, and deep ensembles \cite{gal_dropout_2016,lakshminarayanan_simple_2017}. In supervised learning, uncertainty is often decomposed into aleatoric uncertainty, i.e., inherently random effects (in the data), and epistemic uncertainty, i.e., uncertainty due to a lack of knowledge (about the best model) \cite{gruber_sources_2025,hullermeier_aleatoric_2021,kendall_what_2017}. While existing approaches often provide meaningful uncertainty estimates, they typically require repeated stochastic inference or multiple independently trained models, making them computationally expensive for large-scale Earth observation applications.
Evidential deep learning provides an efficient alternative by directly parameterizing higher-order distributions over model outputs. Sensoy et al. \cite{sensoy_evidential_2018} introduced evidential deep learning for classification, while Amini et al. \cite{amini_deep_2020} extended this framework to regression through DER. DER enables non-Bayesian neural networks to jointly predict continuous targets and associated aleatoric and epistemic uncertainty estimates in a single forward pass. Recent work by Meinert et al. \cite{meinert_unreasonable_2023} demonstrates that DER provides surprisingly competitive uncertainty estimates across a range of regression tasks, highlighting its potential as a computationally efficient alternative to Bayesian approaches.

Recent advances in Earth observation have enabled large-scale mapping of forest structural attributes using multimodal satellite data. In particular, the combination of Sentinel-1 SAR, Sentinel-2 optical imagery, and GEDI LiDAR measurements has substantially improved canopy height estimation at high spatial resolution \cite{kacic_forest_2023,lang_global_2022,potapov_mapping_2021}. While early approaches often relied on classical machine learning models such as random forests \cite{kacic_forest_2023}, deep learning architectures have become the dominant paradigm for forest height estimation. For example, Schwartz et al. \cite{schwartz_high-resolution_2024} and Su et al. \cite{su_canopy_2025} employ multimodal U-Net-based encoder-decoder architectures for canopy height prediction from Sentinel-1 and Sentinel-2 imagery, while Chen et al. \cite{chen_multimodal_2025} demonstrate the benefits of multimodal feature fusion approaches combining multiple Earth observation modalities.

Beyond accurate point prediction, uncertainty-aware forest structure modeling has recently received increasing attention. In particular, Lang et al. \cite{lang_high-resolution_2023} investigated canopy height regression and uncertainty estimation from GEDI waveforms using deep ensembles, highlighting the value of reliable uncertainty estimates for large-scale forest monitoring. Nevertheless, comparatively few studies address uncertainty-aware forest height estimation from multimodal Earth observation data. In contrast, our work focuses on uncertainty-aware forest height estimation by integrating DER into a U-Net-based multimodal Earth observation framework evaluated on the recently introduced TreeUQ benchmark.

\section{Methodology}
\subsection{Problem Formulation}
We evaluate the proposed approach on the TreeUQ benchmark for multimodal forest structure estimation. TreeUQ combines multi-seasonal Sentinel-1 SAR and Sentinel-2 optical imagery with sparse tree inventory annotations across Bavaria at 10 m spatial resolution. Following the original benchmark setup, we use stacked Sentinel-1 and Sentinel-2 composites as multimodal inputs and mean tree height as the regression target. 
More specifically, let $\bfx \in \mathbb{R}^{H \times W \times C}$ denote a multimodal input patch consisting of data from Sentinel-1/-2, where $H$ and $W$ represent the height and width (in pixels) of an input patch ($H = W = 128$) and $C=49$ the number of spectral and categorical input channels. 
The corresponding target $\bfy \in \mathbb{R}^{H \times W}$ contains the mean tree heights of the 10 m pixels in the patch derived from sparse tree inventory measurements provided by the TreeUQ dataset. 
Throughout the remainder of this work, bold symbols denote patch-level quantities, whereas $x$ and $y$ refer to individual pixels within a patch.
The objective is to learn a model that, given an input patch $\bfx$, predicts both the expected tree height and associated predictive uncertainty estimates for each pixel. 
Moreover, in addition to the mean tree height labels, TreeUQ also provides local variance statistics derived from the underlying tree inventory data. In particular, the dataset includes per-pixel tree height variance estimates which are given by the variance of the tree heights between the trees in a pixel.
These statistics enable the assessment of predicted uncertainty estimates and allow us to analyze the relationship between the uncertainty estimates and local structural variability.

\subsection{Deep Evidential Regression for Sparse Supervision}\label{subsec:DER}
DER estimates predictive uncertainty in a single forward pass by parameterizing a higher-order evidential distribution over regression outputs. In accordance with Amini et al. \cite{amini_deep_2020}, the regression targets $y$ are modeled using a Gaussian likelihood with unknown mean and variance, i.e., $y \sim \mathcal{N}(\mu, \sigma^2)$. Following a Bayesian approach, the Normal-Inverse-Gamma (NIG) distribution, $\mathrm{NIG}(\mu, \sigma | m)$, is used as a conjugate prior for the unknown Gaussian parameters $\mu$ and $\sigma$, where $m = (\gamma, \nu, \alpha, \beta)$ are the hyperparameters of the NIG-distribution. Marginalizing over the Gaussian likelihood parameters $\mu$ and $\sigma$, Bayesian inference yields a Student-t predictive distribution for $y$ given $m = (\gamma, \nu, \alpha, \beta)$ with
\begin{align*}
    p(y\mid m)= \frac{p(y | \mu, \sigma, m) p(\mu, \sigma | m)}{p(\mu, \sigma | y, m)} = \mathrm{St}\left(y;\gamma,\frac{\beta(1+\nu)}{\alpha\nu},2\alpha\right).
\end{align*}
Now, if $m$ is known, the prediction of $y$ is given by $\E[\mu] = \gamma$ and aleatoric and epistemic uncertainty can be calculated as
\begin{align*}
    u^2_{al}=\frac{\beta}{\alpha-1}; \qquad 
    u^2_{ep}=\frac{\beta}{\nu(\alpha-1)}
\end{align*}
with total predictive uncertainty $u^2_{pred} = u^2_{al} + u^2_{ep}$.
Finally, the idea of DER is to train a neural network to learn the parameters $m$ from data with the evidential loss $\mathcal{L}_\mathrm{DER}$ proposed by Amini et al. \cite{amini_deep_2020} as the objective, consisting of a negative log-likelihood term and an evidence regularizer that penalizes overconfident incorrect predictions.

The formulation above follows the original DER framework \cite{amini_deep_2020}. To adapt DER to the sparse supervision setting of TreeUQ, we introduce the following masked evidential loss.
In particular, we define a binary patch-wise validity mask $\bfdelta \in \{0,1\}^{H \times W}$, where $\delta_{ij}=1$ indicates that a valid tree height label is available at spatial location $(i,j)$ within the patch. 
Then, during training, loss computation is restricted to valid pixels only.
More precisely, the masked evidential loss for a patch is computed as 
\begin{align}\label{eq:loss}
    \mathcal{L}_\mathrm{mDER}(\bfw; \bfy, \bfm) = \frac{1}{\sum_{i,j} \delta_{ij}} \sum_{i,j} \delta_{ij} \cdot \mathcal{L}_\mathrm{DER}(\bfw; y_{ij}, m_{ij}),
\end{align}
where $m_{ij}$ and $y_{ij}$ denote the pixel-wise predicted evidential parameters and the pixel-wise targets at location $(i,j)$, respectively, $\bfm \in \R^{H \times W \times 4}$ denotes the evidential parameters for the patch and $\bfw$ the neural network weights. This masking strategy prevents missing labels from influencing uncertainty estimation and enables stable optimization under sparse geospatial supervision.

\subsection{Model architecture}
In order to assess the predictive performance of the DER-approach, two different model architectures are trained: One U-Net is trained to output a point prediction for the mean tree height and one is trained with an evidential head to output the evidential parameters.
Both models share the same U-Net architecture \cite{ronneberger_u-net_2015} with a ResNet-50 encoder pre-trained on ImageNet \cite{he_deep_2016} and adapted to the 49 input channels of TreeUQ.
The deterministic U-Net predicts a single tree height value per pixel and is trained using a masked Smooth $L_1$ loss. In contrast, the evidential U-Net replaces the scalar regression head with an evidential output layer of four output neurons $(z_\gamma, z_\nu, z_\alpha, z_\beta)$ that predicts the NIG parameters $m=(\gamma,\nu,\alpha,\beta)$, where $\gamma=z_\gamma$, $\nu=\mathrm{softplus}(z_\nu)$, $\alpha=\mathrm{softplus}(z_\alpha)+1$, and $\beta=\mathrm{softplus}(z_\beta)$. Training is performed using the masked evidential loss defined in Equation \eqref{eq:loss}, and the predicted mean tree height for an input $x$ is given by its predicted value for $\gamma$.
Apart from the output parameterization and loss function, both models are identical, providing a controlled comparison between deterministic and uncertainty-aware regression.

\section{Experimental Setup}

Following the official TreeUQ benchmark, geographically disjoint train, validation, and test regions are used to prevent spatial leakage and to evaluate generalization to unseen geographic areas. Since tree heights exhibit a strongly right-skewed distribution, the benchmark applies a $\log(1+y)$-transformation to stabilize optimization. 
Both model configurations (i.e., with and without evidential head) share identical preprocessing, masking, optimization, and evaluation settings. Each model is trained independently using eight random seeds while keeping all hyperparameters fixed. Models are optimized using AdamW with a learning rate of $10^{-4}$, a weight decay of $10^{-4}$, and a cosine annealing learning rate schedule. Moreover, mixed-precision training is employed throughout all experiments, and a batch size of 32 is used.
Data augmentation is applied exclusively to the training dataset and consists of random horizontal and vertical flips ($p=0.5$) as well as random $90^\circ$ rotations ($p=0.75$). All augmentations are applied jointly to the input data, target maps, and validity masks to preserve spatial correspondence.
Validation is performed every five epochs. Then, the model checkpoint achieving the lowest validation RMSE is retained for evaluation. Finally, training is terminated early if no improvement in validation RMSE is observed for ten consecutive validation steps, with a maximum training budget of 100 epochs.

After training, each seed is evaluated on the held-out test
split without augmentation.
We report masked RMSE, MAE, and $R^2$ in original units and in the
$\log(1+y)$-transformed space, aggregating over valid test pixels only.

\section{Results}
Table \ref{tab:prediction_accuracy} summarizes the predictive performance of the deterministic and evidential U-Net models on the held-out test set. While the deterministic U-Net achieves slightly lower errors and higher $R^2$, the performance differences are small relative to the observed variability across random seeds. In particular, the DER U-Net attains an RMSE of $5.91 \pm 0.15$m compared to $5.75 \pm 0.09$m for the deterministic baseline. These results indicate that incorporating evidential uncertainty estimation does not substantially degrade predictive performance while additionally providing uncertainty estimates in a single forward pass.
\begin{table}
\caption{Prediction accuracy of deterministic U-Net and evidential U-Net.}
\label{tab:prediction_accuracy}
\centering
\begin{tabular}{l|r|r|r}
\hline
Model & RMSE (m) & MAE (m) & $R^2$ \\
\hline
U-Net     & $5.75 \pm 0.09$ & $4.40 \pm 0.06$ & $0.525 \pm 0.015$ \\
DER U-Net & $5.91 \pm 0.15$ & $4.53 \pm 0.15$ & $0.497 \pm 0.026$ \\
\hline
\end{tabular}
\end{table}

\begin{figure}
    \centering
    \includegraphics[width=0.4\linewidth]{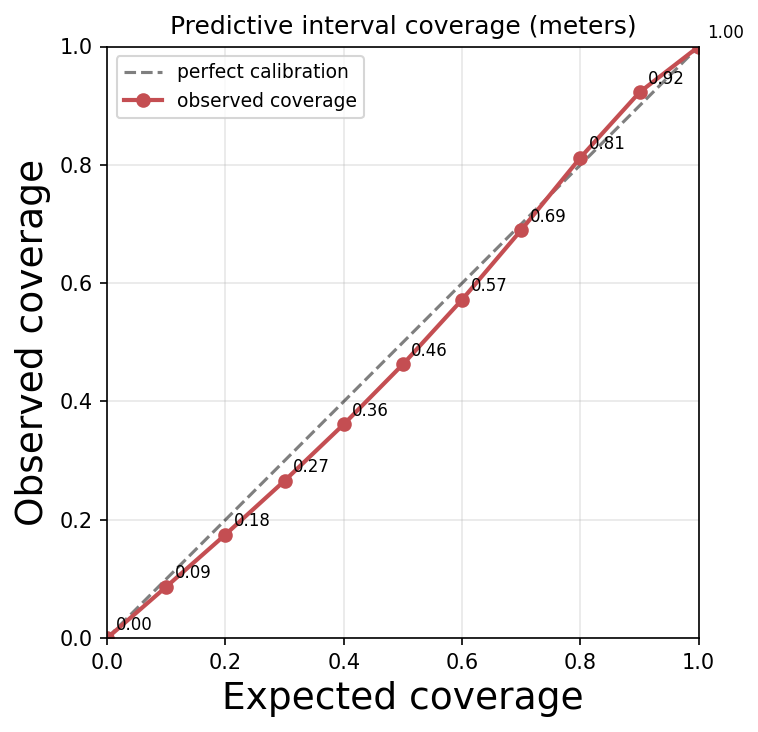}
    \caption{Calibration curve for the DER U-Net. The ideal calibration would follow the diagonal. The ECE for the model is 0.025.}
    \label{fig:coverage_curve}
\end{figure}

To assess the quality of the predicted uncertainty estimates of the evidential U-Net, we evaluate calibration using prediction interval coverage. Following Kuleshov et al. \cite{kuleshov_accurate_2018}, for each observation, prediction intervals are constructed from the predictive mean and total predictive uncertainty for various confidence levels, and empirical coverage is eventually compared to the nominal confidence level for the different coverage levels. Consequently, the observed coverage ideally matches the expected coverage and the ideal curve is given by the diagonal.
The calibration curve shown in Figure \ref{fig:coverage_curve} closely follows the ideal diagonal, indicating that the predicted uncertainties are generally well calibrated. This is supported by a low Expected Calibration Error (ECE) of 0.025, corresponding to an average deviation of 2.5 percentage points between nominal and empirical coverage. Minor deviations from perfect calibration can nevertheless be observed: the model tends to be slightly overconfident for coverage levels below 70\%, while exhibiting mild underconfidence for larger prediction intervals. Overall, the results suggest that the uncertainty estimates produced by the DER model provide a reliable characterization of predictive uncertainty.

\begin{figure}
    \centering
    \includegraphics[width=0.8\linewidth]{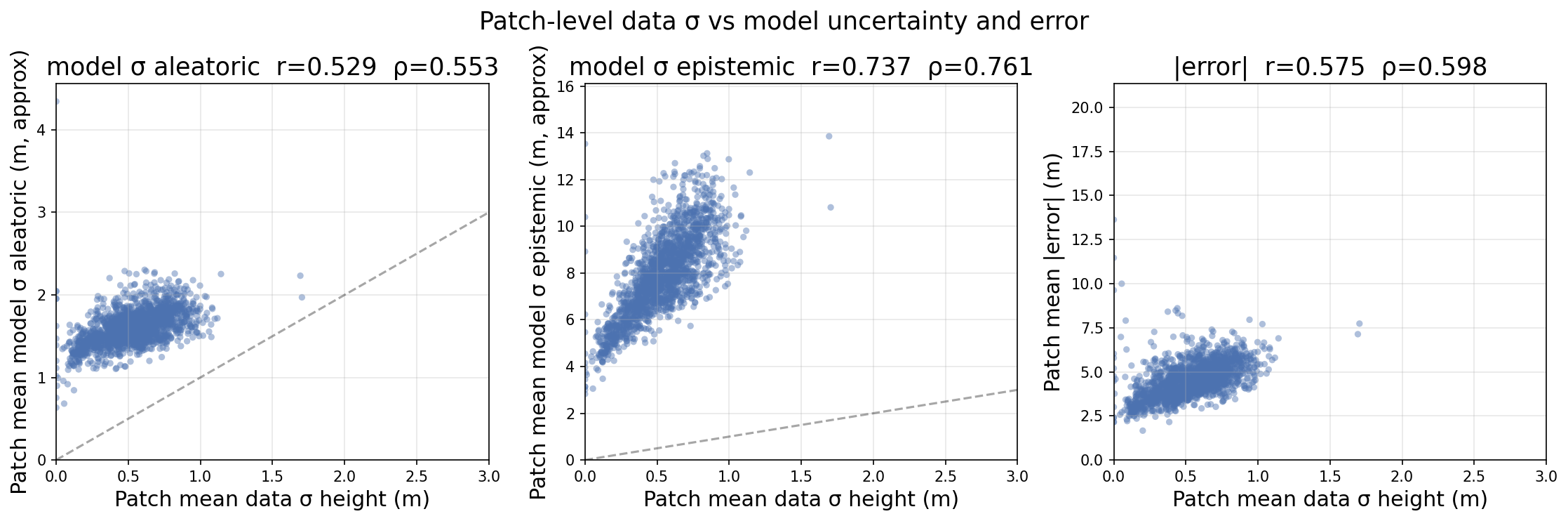}
    \caption{Correlation of the patch-wise tree height variances in TreeUQ with the aleatoric uncertainty estimates, the epistemic uncertainty estimates and the prediction errors, respectively.}
    \label{fig:correlation_variance_aleatoric}
\end{figure}

Beyond calibration, TreeUQ provides local tree-height variance statistics derived from the underlying inventory measurements. However, these variance estimates should not be interpreted as ground-truth aleatoric uncertainty. While the predicted aleatoric uncertainty quantifies the expected variability of a tree-height prediction given the available imagery, the variance provided in the dataset measures the variability of individual tree heights within a 10 m pixel. The two quantities therefore characterize different sources of variability.
Nevertheless, a positive relationship can be expected. Regions containing a heterogeneous forest structure and a wider range of tree heights are likely to be more difficult to model from remote sensing data, which should be reflected in both increased predictive uncertainty and larger prediction errors. To investigate this hypothesis, we compare the inventory-derived variance statistics with the predicted aleatoric and epistemic uncertainty estimates and the absolute prediction error.
At the individual-pixel level, no meaningful relationship is observed. We therefore aggregate all quantities at the patch level. This aggregation reduces local sampling noise and better reflects forest structural heterogeneity. Furthermore, uncertainty estimates produced by the U-Net are influenced by spatial context through the model’s receptive field, making patch-level comparisons more appropriate. After aggregation to the patch level, clear positive associations are visible (see Figure \ref{fig:correlation_variance_aleatoric}).

\vspace{-0.2cm}
In particular, the inventory-derived variance exhibits a moderate correlation with the predicted aleatoric uncertainty (Pearson $r=0.53$, Spearman $\rho=0.55$). Interestingly, an even stronger relationship is observed for epistemic uncertainty ($r=0.74$, $\rho=0.76$). This indicates that structurally heterogeneous forest stands are not only associated with intrinsic variability in the target variable, but also with increased model uncertainty, potentially reflecting regions that are more difficult to represent from the available satellite observations. Moreover, patch-wise prediction errors increase with the inventory-derived variance ($r=0.58$, $\rho=0.60$), suggesting that regions with larger within-pixel tree-height variability are inherently more difficult to predict from satellite observations.
Taken together, these findings indicate that the uncertainty estimates produced by the evidential U-Net are aligned with meaningful characteristics of the underlying forest structure.

\begin{figure}
    \centering
    \includegraphics[width=1\linewidth]{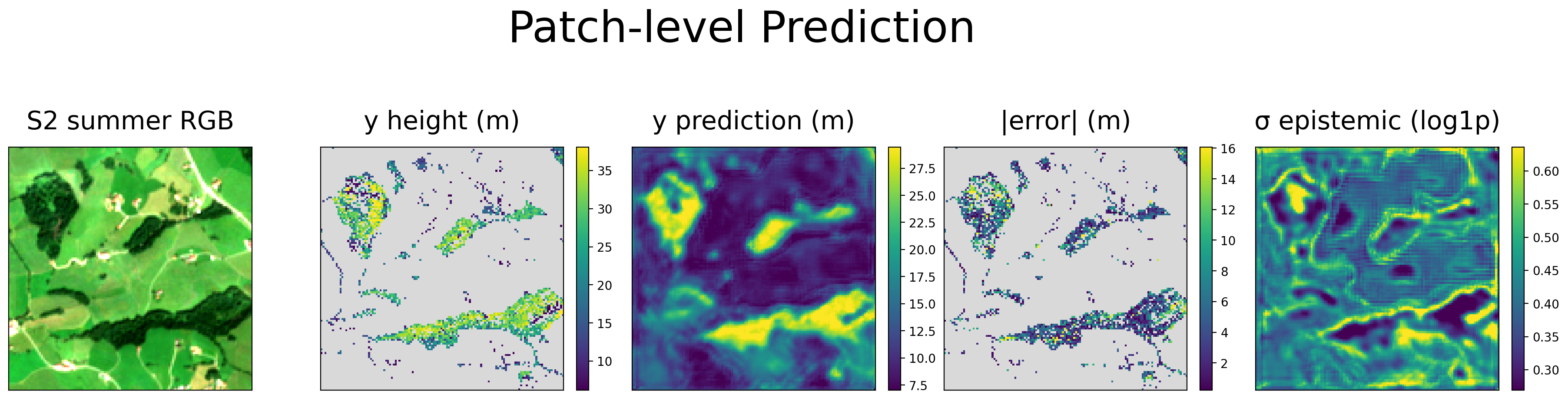}
    \caption{Original RGB photo, true mean height, prediction, prediction error and predicted total uncertainty for a representative patch in the test data.}
    \label{fig:prediction_map}
\end{figure}

Finally, Figure \ref{fig:prediction_map} illustrates the spatial distribution of predictive uncertainty for a representative test patch. Higher epistemic uncertainty is observed near transitions between forested and non-forested areas, whereas uncertainty is generally lower within homogeneous forest stands. These patterns indicate that the model appropriately expresses reduced confidence in regions that are more difficult to predict. The figure also highlights a limitation of the sparse supervision setting: since the masked training loss is evaluated only on pixels with valid tree inventory annotations, the model is never explicitly trained to predict zero height in non-forest regions. Consequently, both the deterministic and evidential models occasionally produce spurious tree-height predictions in background areas.
\vspace{-0.3cm}

\section{Discussion and conclusion}
In this work, we investigated DER for uncertainty-aware forest height estimation from multimodal Earth observation data. Using a U-Net with an evidential output head, the proposed approach jointly predicts mean tree height along with aleatoric and epistemic uncertainty estimates per pixel in a single forward pass. To account for the extreme label sparsity of the TreeUQ benchmark, we introduced a masked evidential loss that restricts training to pixels with valid tree inventory annotations.
The results show that the evidential model achieves predictive performance comparable to that of a deterministic U-Net while additionally providing meaningful and well-calibrated uncertainty estimates. Qualitative analyses further suggest that uncertainty increases in challenging regions, such as transitions between forested and non-forested areas.
However, several limitations should be noted. While DER provides a decomposition into aleatoric and epistemic uncertainty, the interpretation of these quantities remains heuristic and should be treated with caution 
\cite{meinert_unreasonable_2023}. Moreover, the evidential framework relies on a Gaussian likelihood assumption, which may not fully capture the distribution of forest heights despite the applied log-transformation. Finally, due to the sparse supervision setting, the model is never explicitly trained to predict zero heights in non-forest regions. As a result, both models may produce spurious height predictions in areas where no trees are present.
Overall, our findings suggest that DER constitutes a simple and computationally efficient framework for uncertainty-aware forest structure estimation under sparse geospatial supervision. Future work should compare DER with alternative uncertainty estimation approaches, and extend the analysis to additional forest structure variables and Earth observation benchmarks.

\begin{credits}
\subsubsection{\ackname} The present contribution is supported by the Helmholtz Association under the joint research school “HIDSS-006 - Munich School for Data Science@Helmholtz, TUM \& LMU.
\end{credits}
%
%

\bibliographystyle{splncs04}
\bibliography{references}
%




\end{document}